\documentclass[graybox]{svmult}

\usepackage{type1cm}        
\usepackage{makeidx}         
\usepackage{graphicx}        
\usepackage{multicol}        
\usepackage[bottom]{footmisc}

\usepackage{newtxtext}       %
\usepackage{newtxmath}       

\usepackage{subcaption}
\usepackage[mathscr]{euscript}
\usepackage{dsfont}
\usepackage{amsmath}
\usepackage{bm}  
\usepackage{array}

\makeindex             

\begin{document}

\title*{Style Similarity as Feedback for Product Design}

\author{Mathew Schwartz, Tomer Weiss, Esra Ataer-Cansizoglu, and Jae-Woo Choi}
\authorrunning{Schwartz, Weiss, Cansizoglu, and Choi}

\institute{Mathew Schwartz \at New Jersey Institute of Technology, Newark,NJ, \email{cadop@umich.edu}
\and Tomer Weiss \at New Jersey Institute of Technology, Newark NJ, \email{tweiss@njit.edu}
\and Esra Ataer-Cansizoglu \at Facebook, Boston MA, \email{cansizoglu@ieee.org}
\and Jae-Woo Choi \at Wayfair, Boston MA, \email{jchoi@wayfair.com}
}
%
%
\maketitle

\abstract*{Matching and recommending products is beneficial for both customers and companies. With the rapid increase in home goods e-commerce, there is an increasing demand for quantitative methods for providing such recommendations for millions of products. This approach is facilitated largely by online stores such as Amazon and Wayfair, in which the goal is to maximize overall sales. Instead of focusing on overall sales, we take a product design perspective, by employing big-data analysis for determining the design qualities of a highly recommended product. Specifically, we focus on the visual style compatibility of such products. We build off previous work which implemented a style-based similarity metric for thousands of furniture products. Using analysis and visualization, we extract attributes of furniture products that are highly compatible style-wise. We propose a \textit{designer in-the-loop} workflow that mirrors methods of displaying similar products to consumers browsing e-commerce websites. Our findings are useful when designing new products, since they provide insight regarding what furniture will be strongly compatible across multiple styles, and hence, more likely to be recommended.}

\abstract{Matching and recommending products is beneficial for both customers and companies. With the rapid increase in home goods e-commerce, there is an increasing demand for quantitative methods for providing such recommendations for millions of products. This approach is facilitated largely by online stores such as Amazon and Wayfair, in which the goal is to maximize overall sales. Instead of focusing on overall sales, we take a product design perspective, by employing big-data analysis for determining the design qualities of a highly recommended product. Specifically, we focus on the visual style compatibility of such products. We build off previous work which implemented a style-based similarity metric for thousands of furniture products. Using analysis and visualization, we extract attributes of furniture products that are highly compatible style-wise. We propose a \textit{designer in-the-loop} workflow that mirrors methods of displaying similar products to consumers browsing e-commerce websites. Our findings are useful when designing new products, since they provide insight regarding what furniture will be strongly compatible across multiple styles, and hence, more likely to be recommended.}

\section{Introduction}\label{sec:intro}

\begin{figure}[h]
\includegraphics[width = \textwidth, page=2]{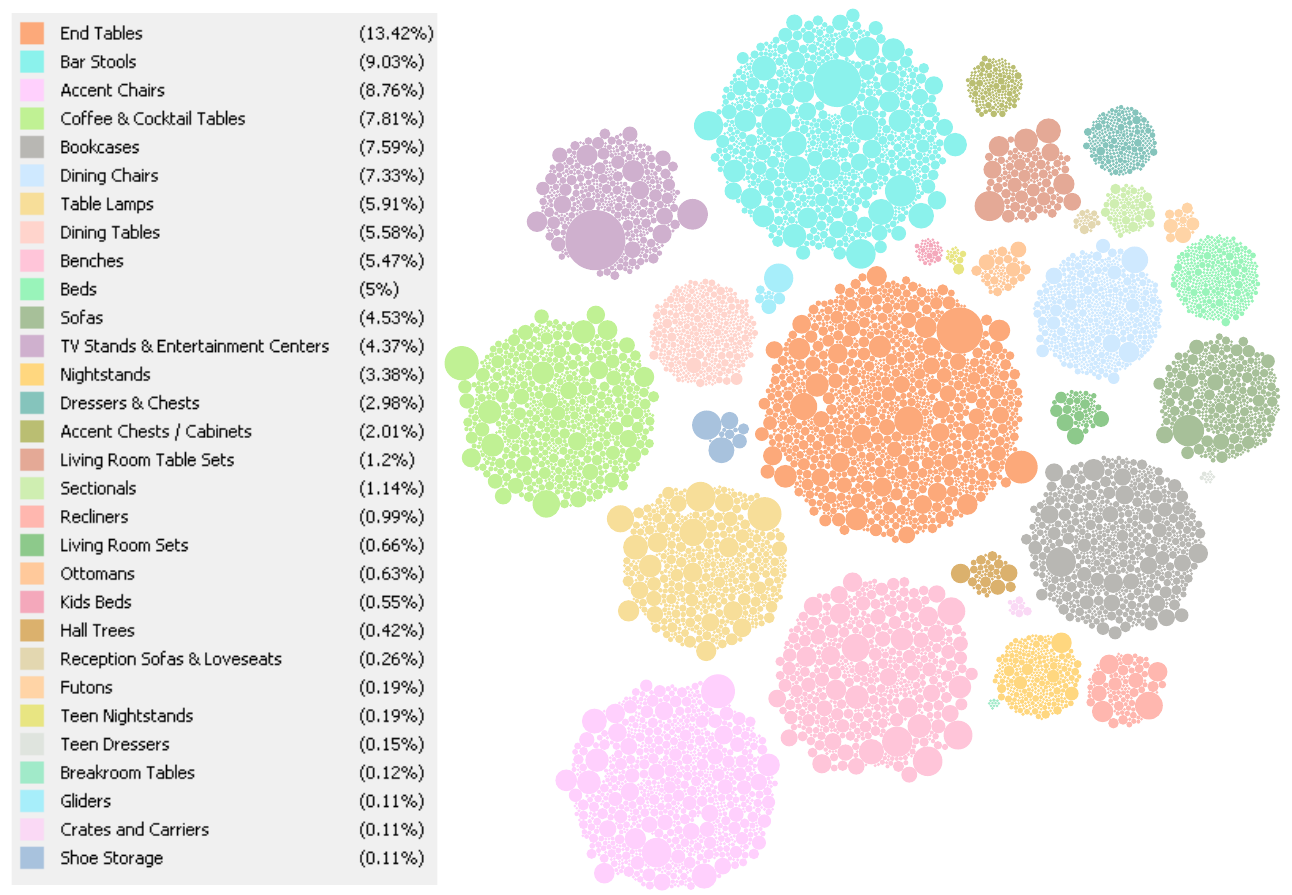}
\caption{The nodes of the reduced dataset. Color corresponds to the product group type and size corresponds to the degree of the node. }
\label{fig:sku_nodes_groups}       
\end{figure}

Interior designers and architects are experienced in creating cohesive styles of an interior space. However, the cost of such a professional is prohibitive for most people. Such professionals use intuition, experience, rules in composition, and color theory to drive decision making, along with modes of visual communication such as drawing and 3D rendering. For the typical consumer, interior design and home decoration are a balance of personal taste and guesswork. One of the more difficult barriers to overcome on the general consumer side is the visualization of multiple products within their context (i.e., a room all products will reside in). Similarly, the sorting of massive online catalogs of products and matching these products in a desirable way can be overwhelming. 

One sorting approach is through using a set of attributes and labels assigned to products in the catalog. Through such attributes, item-based filtering can be applied, in which items sharing certain features, such as color, are recommended to a user. While this technique is a standard industry practice, it is limited in usefulness for large datasets containing hundreds of thousands of products. Additionally, certain attributes are more abstract and difficult to ascertain, requiring multiple ratings and labels by experts--some of which might not agree on such subjective matter. One such attribute is the style of a given space, which includes multiple components such as: fabric, color scheme, material, furniture style and flooring. In this work, we use such components as the basis of 4 major room styles:\emph{modern, traditional, cottage, coastal},  that are popular within an interior product e-commerce website (Wayfair). We elaborate on descriptions of these styles in Table~\ref{tbl:style_descriptions}. 

An emerging trend in the e-commerce space is to use image-based analysis for identifying similarities of products. This is useful in online situations where products must have an image (or 3D rendering) for customer display, while not being fully labelled with information describing subjective style-related attributes. In this paper we propose a synergy between the methods used for recommendation algorithms to a consumer and the design of the products to be recommended. Background and implementation of past work is presented, along with an in-depth discussion on the resulting analysis of our dataset. This dataset is of images labeled with product Stock Keeping Units (SKUs) and similarly, associated with a product group (Fig.~\ref{fig:sku_nodes_groups}) from the Wayfair company. By organizing data into a network structure, we are able to visualize and explore connectivity and relationships between various attributes and types. Finally, we present a flowchart for how such image-based analysis of style similarities can be integrated within a product designers workflow, in a not too dissimilar way to that of a consumer browsing online. 

\section{Related Work}

Understanding a user's preferences lies at the heart of many e-commerce websites~\cite{ataer2019room,sachidanandan2019designer,jing2015visual}. An accurate representation of customer's style preferences--enables better product recommendations, and a more personalized shopping experience. The gap in matching an individuals preference with a product is in the extraction and categorization of style-related attributes of a product. Researchers have proposed multiple quantitative methods for understanding (i.e., extracting) subjective attributes related to aesthetics and fashionablity that compose a style~\cite{Hsiao2017, Schifanella2015, Dhar2011, Simo2015, Pan2019, lun2015elements} with a few that focus on interiors~\cite{Pan2019, lun2015elements, weiss2018fast}.

Quantitative methods for style recognition are typically data-driven and employ a variety of machine learning techniques. Recently, neural networks, and specifically deep neural networks, have become paramount for automatically assessing images~\cite{goodfellow2016deep} and in general are effective for image recognition tasks~\cite{simonyan2014very}. A large body of work exists on the categorization of images using such deep neural networks. To find an image's category, a deep neural network learns a lower dimensional representation, which we refer to as style embeddings (See Sec.~\ref{sec:learning}), based on labeled data. These embeddings represents an image's category in an abstract category space, such as weightings within various style types. Distances between embeddings representing different images can then be evaluated to find whether they are similar~\cite{hoffer2015deep}. The two most frequently used data types used by researchers to learn style and visual compatibility are images and/or 3D models.

Using 3D models, Hu et al.~\cite{hu2017style} presents a method for discovering elements that characterize a given style, where the elements are co-located across different 3D models associated with such style. Both Liu and Lun et al.~\cite{liu2015style,lun2015elements} propose to learn 3D model style based on perceptual geometric elements. They employ crowd-sourcing to quantify the different geometric style components. Building on this work, Lim and colleagues~\cite{lim2016identifying} use neural networks to directly identify the style of 3D models, without geometric features, by extracting images of the target 3D model. While the Wayfair e-commerce data consists of 3D models, we focus on the more general use-case of image data for determining style-similarity. Using images, Karayev et al.~\cite{BMVC.28.122} learn the style of paintings and common photographs using linear classifiers. Thomas and Kovashka~\cite{thomas2016seeing} use style to identify a photograph's artist, which could have implications for brand-identity. Considering a products connection to other products and within societal context, Simo-Serra and colleagues~\cite{simo2016fashion} propose to predict an outfits' fashionabilty using neural networks and Gu et al.~\cite{gu2017understanding} identify fashion trends from street photos. Similar to our work, but focusing directly on product images rather than overall style,~\cite{bell2015learning} used deep networks for a visual search system that can aid in interior design through product image similarity.

For design-based companies, understanding consumer interests and brand recognition is strongly linked to visual style attributes. Studies into quantifying design aesthetics for consumer decisions have provided valuable insights through eye-tracking and interactive modeling~\cite{hyun2017gap}. As elaborated by Burnap and colleagues~\cite{burnap2016balancing}, the importance of brand recognition, and the difficulty in balancing with design freedom, is imperative for companies in brand-conscious spaces. While that work was within the automotive industry, cohesion between various single products for a complete style in an interior space is equally important. This cohesion among products as seen by a consumer should be achieved on both the e-commerce side--through intuitive and appropriate recommendations to a consumer; as well as the product design side--by varied designs useable in both different multiple product groupings (a range of styles) and along with other products (including with competing brands).

\begin{table}[!t]
\caption{Descriptions of $4$ major room styles: modern, traditional, cottage, coastal. Major style features include fabric, color, material, furniture, and flooring.}
\label{tbl:style_descriptions}       
%
%
\begin{tabular}{p{.12\textwidth}>{\raggedright}p{.22\textwidth}>{\raggedright}p{.22\textwidth}>{\raggedright}p{.22\textwidth}>{\raggedright}p{.22\textwidth}}
\hline\noalign{\smallskip}
&\textbf{Modern} & \textbf{Traditional} & \textbf{Cottage} & \textbf{Coastal} \tabularnewline
\noalign{\smallskip}\svhline\noalign{\smallskip}
    \textbf{Fabric} & heavy texture, leathers, linens & damask or jacquard, velvet or silk, chintz or florals & soft florals, linen, checks and gingham, toile & linen, stripes, nautical \tabularnewline
    
    \textbf{Color Scheme} & muted solids in neutrals, greys and blacks & blue, dark red, hunter green and brown & muted blues, pinks, reds and greens, white, pale yellows, soft greens & blue, white, red, green\tabularnewline
    
    \textbf{Furniture} & sleek, low to the ground, clean lines, straight legs on base & dark wood, gold accents, antique & slightly distressed, vintage inspired,skirted sofas or chairs, feminine accents, wooden signs &  whitewashed, distressed, beadboard accents, bamboo and rattan\tabularnewline
    
    \textbf{Material} & mixed & marble, gold, cherry or mahogany wood & white washed or cherry wood, straw baskets and worn metals & reclaimed or painted wood, seeded glass or beach glass, beach wood\tabularnewline
    
    \textbf{Flooring} & stripes or natural fiber rugs such as jute or sisal & ornately patterned carpets & braided cotton, soft floral or checked rugs & stripes or woven, seascape prints, sisal or jute\tabularnewline
\noalign{\smallskip}\hline\noalign{\smallskip}
\end{tabular}

\end{table}

\section{Learning Image Style with Neural Networks}
\label{sec:learning}
Our work relies on deep neural networks to learn furniture styles through images.  Training images were collected in a previous study~\cite{ataer2019room, ilkay2020} and each was labeled with a designated style by multiple experts (Section \ref{sec:labeling}). Our neural network is trained to estimate an approximate style of an image, and in doing so, provides a comparative stylistic differences between the images (Section~\ref{sec:nn}). This stylistic difference is determined by a 16-dimensional layer in the neural network which we denote as \emph{style embeddings}. Given such embeddings, we are able to retrieve similar style images (Section~\ref{sec:eval}), and by doing so, recommend similar-style products.
For a more detailed explanation than the following overview, please refer to the previous work, done in conjunction with Wayfair's Data Science team~\cite{ataer2019room,ilkay2020}.

\begin{figure}[b!]
\captionsetup[subfigure]{labelformat=empty}

\centering
\subcaptionbox{Modern}{\includegraphics[height=2.15cm,keepaspectratio]{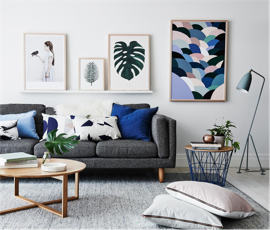}}
\subcaptionbox{Traditional}{\includegraphics[height=2.15cm,keepaspectratio]{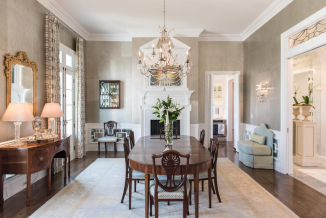}}

\subcaptionbox{Cottage}{\includegraphics[height=2.15cm,keepaspectratio]{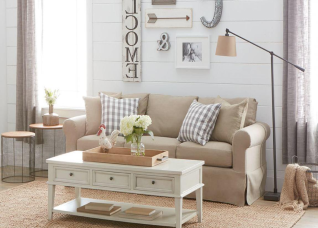}}
\subcaptionbox{Coastal}{\includegraphics[height=2.15cm,keepaspectratio]{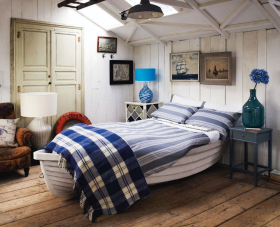}}
\caption{Examples of major room styles.}
\label{fig:style_def}
\end{figure}

\subsection{Labeling Image Style}
\label{sec:labeling}
Our initial dataset contains $672,000$ images of interior scenes and products collected from various sources: 
(i) Staging a scene in a physical room, and then capturing it with a camera, (ii) Captured from a virtual scene, curated by a 3D artist, and (iii) Scraped from 3rd parties. However, due to noisy image metadata, an image's origin source might not always be known. Each image is designated with an interior style label which includes: modern, traditional, cottage, coastal (Figure \ref{fig:style_def}), where each style is described with certain criteria about fabric, color scheme, material, furniture style and flooring (Table~\ref{tbl:style_descriptions}). 

Style labeling is a subjective task prone to noise and variation among multiple people, including stylists. To alleviate discrepancies in dataset labeling, 10 independent experts were used to designate a style label to images. All data, including discrepancies between the expert labels were recorded, leading to a noisy dataset of labels--a common issue when training a neural network--as it is prone to inaccurate estimations. Such challenges were overcome by having the neural network compare images within the same dataset and their attributed labels from multiple experts. This assumes the existence of tangible differences between each label that knowledgeable experts may discern between, similar to the methods described in~\cite{bradley1952rank}.

A new set of comparison labels were generated based on the expert-determined style labels. During training of the neural network a randomly selected sample of images were used in which, for each style, and for each pair of images, a comparison label is generated with respect to the relative order of the number of style labels each image receives. Formally, for a given style and an image pair, a comparison label will be $+1$ if in that style, the first image has $x$ more expert votes than the second image in the pair, and $-1$ if it has less votes. Variations of these values of voting thresholds were experimented with, i.e.,  $x={1,2,...}$ and any image pair was discarded that fell outside of the defined range.

\subsection{Neural Network Configuration}
\label{sec:nn}
We use an architecture inspired by siamese neural networks~\cite{bromley1994signature} that extends the Bradley-Terry~\cite{bradley1952rank}, a neural network that learns from comparisons. VGG16~\cite{simonyan2014very} forms the base of our siamese neural network. The last layer of VGG16 is removed and instead consecutively add two fully connected layers: a 16-dimensional layer, which informs us of an image's style embeddings, and a 4-dimensional layer, which was used to estimate style.

For the model, 80\% of images were allocated for training, 10\% for validation, and 10\% for testing. Thus, images in the training set were not paired with images in the validation or test sets for generating comparison labels. Our style estimation model was trained on comparisons selected uniformly at random over the training dataset. Next, the resulting models were evaluated on the validation set to determine the optimal neural network parameters. This model was implemented with Python and Tensorflow, where training the neural network model took about 5 hours on a NVIDIA Tesla V100 GPU.

\begin{figure}[t]
  \centering
  \includegraphics[width=0.49\textwidth]{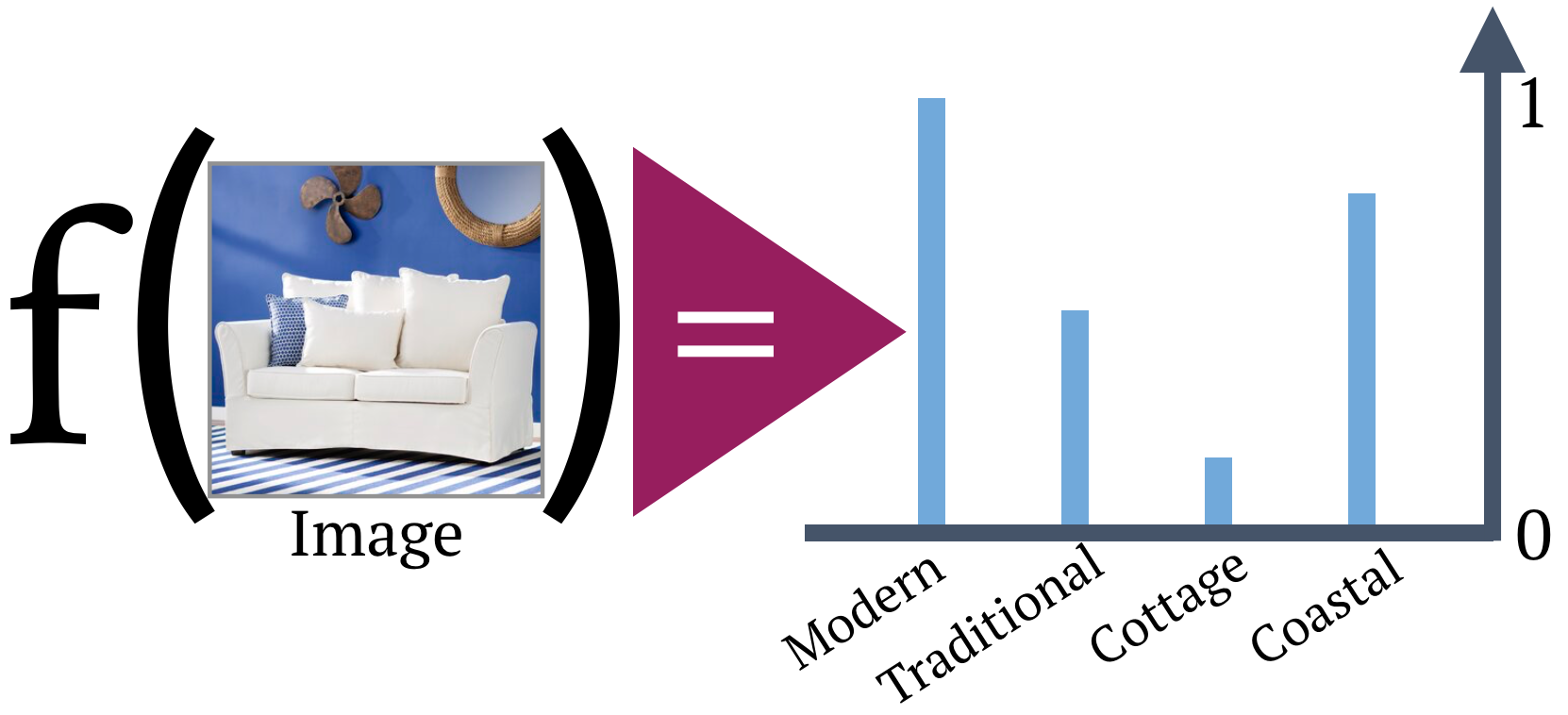}
  \caption{Our neural network model learns furniture style from images. Given a context image with a target furniture piece (left), we predict its style embeddings.}
\label{fig:embeddings}
\end{figure}

\subsection{Style Estimation}
\label{sec:eval}
Since our goal is to accurately estimate an image's style, we preformed several experiments to assess our neural network performance. Each experiment involved random subsets of our image dataset, and different neural network parameter combinations, including thresholding of expert votes for the comparison labels. Such multiple experiments aid in finding the optimal set of parameters that can most accurately estimate the style of a given image.

For estimating an image's style, we use the last layer of the neural network, which has 4 dimensions, each of which provide a probabilistic estimation of an image's style as modern, traditional, cottage, and coastal (Figure~\ref{fig:embeddings}). To evaluate the estimation accuracy, we found the maximum estimation values over all predicted style labels. Our model achieved an average style estimation accuracy of $79\%$ for a random image in our dataset. This estimation accuracy varied by style, with $86.9\%$ estimation accuracy for an image labeled as modern, $74.6\%$ for traditional, $71.6\%$ for cottage, and $67.9\%$ for coastal. One possible explanation to the differences in accuracy is that images with style labels such as coastal can also be subjectively considered as part of another style, such as modern or traditional. Adding more images to our dataset might improve the estimation accuracy per style-class.

\begin{figure}[t!]
\centering
\includegraphics[height=2.08cm,keepaspectratio]{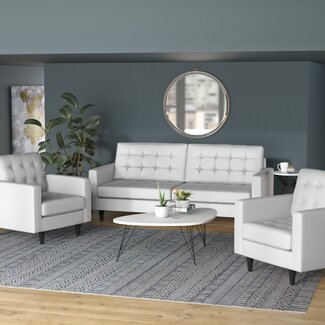}
\includegraphics[height=2.08cm,keepaspectratio]{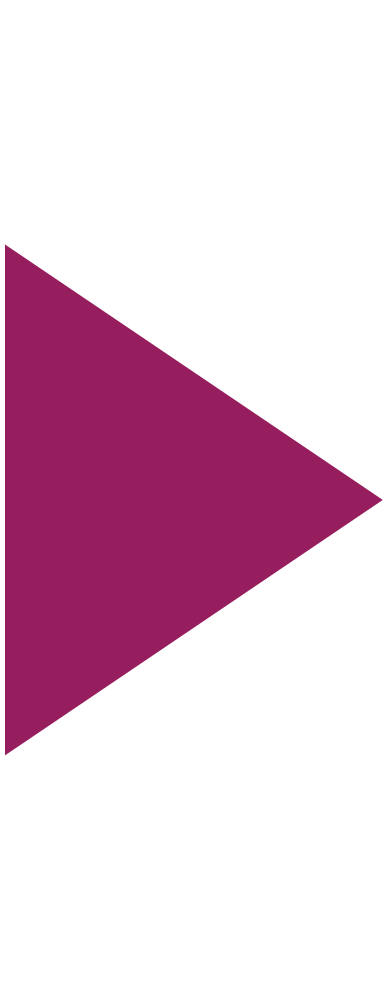}
\includegraphics[height=2.08cm,keepaspectratio]{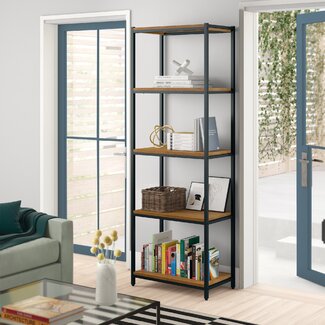}
\includegraphics[height=2.08cm,keepaspectratio]{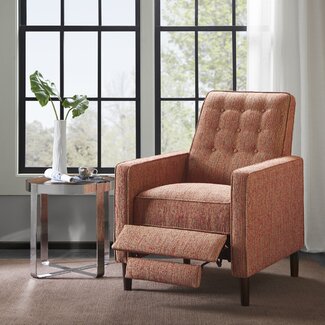}
\includegraphics[height=2.08cm,keepaspectratio]{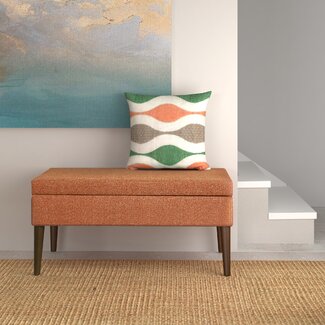}
\includegraphics[height=2.08cm,keepaspectratio]{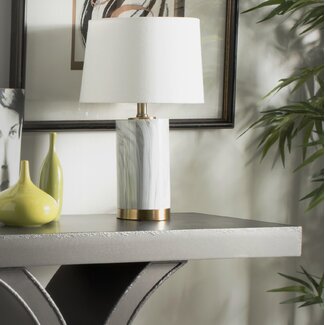}
\caption{Product images used for recommendations. Given the furniture image on the left, our model suggests the recommendations on the right.}
\label{fig:product_examples}
\end{figure}

For recommending similar style images, we use the second to the last layer, which includes 16 dimensions. Given a seed image, we consider this layer as the style embeddings of the image. To evaluate our framework's ability to recommend similar style images, we retrieved the nearest image(s) by using the Euclidean distance from the seed image's embeddings. Figure~\ref{fig:product_examples} demonstrates 
the $5$ closest images to a given seed image, ranked from left to right, ordered by the distance to the seed image. To measure our recommendation performance, we checked if the style labels of the retrieved images match the style of the seed image. Our network's retrieval performance is $73.9\%$ on an randomly sampled image, which is a significant improvement over previous work~\cite{ataer2019room}.

\section{Analysis of Style for Product Design}
\subsection{Data Processing}
We use Networkx~\cite{hagberg2008exploring} in python for constructing a graph from the relationship data. This graph is then imported to Gephi~\cite{ICWSM09154} for visualization. We refer to a given SKU, or SKU and its subsequent options--as a product. 
By taking the resulting style embeddings comparison for the images of the product dataset from the Wayfair e-commerce site, the graph is constructed by associating products as nodes and the euclidean distance between image style embeddings (Section \ref{sec:eval}) as weights.  As the distance between images is bi-directional, the result is a weighted undirected graph. To visually analyze such relationships, edge weights are recalculated to $\frac{1}{dist}$, so that a higher weight corresponds to stylistically stronger similarity. As our data is based on raw e-commerce listings, the staging and advertising of products overlaps between images. To reduce strong connections between images in which the same products exist albeit in a different setting, the edges within the graph of overlapping products (when this is known) are removed. 

For a more fine-grained analysis of our dataset, we needed to further reduce the resulting graph. After duplicate data processing, a total of $17,819$ products and $2,485,878$ weights existed in the graph, with weights ranging from 0.001 to over 100000, the latter in which images are nearly identical. To remove such noise from the dataset, we cut the graph to include only edges with a weight between 1 and 10. Next, we removed products that were in a group category with less than 10 other products. The remaining graph is composed of $8,947$ nodes and $33,154$ edges. Important to keep in mind, the model we use for this paper was designed for style-similarity, which as pointed out in~\cite{bell2015learning}, is different than image similarity as an image can be visually similar and hence likely stylistically similar, but similar styles may be different visually.

\subsection{Product Relationships}
\begin{figure}[h]
\includegraphics[width = \textwidth, page=1]{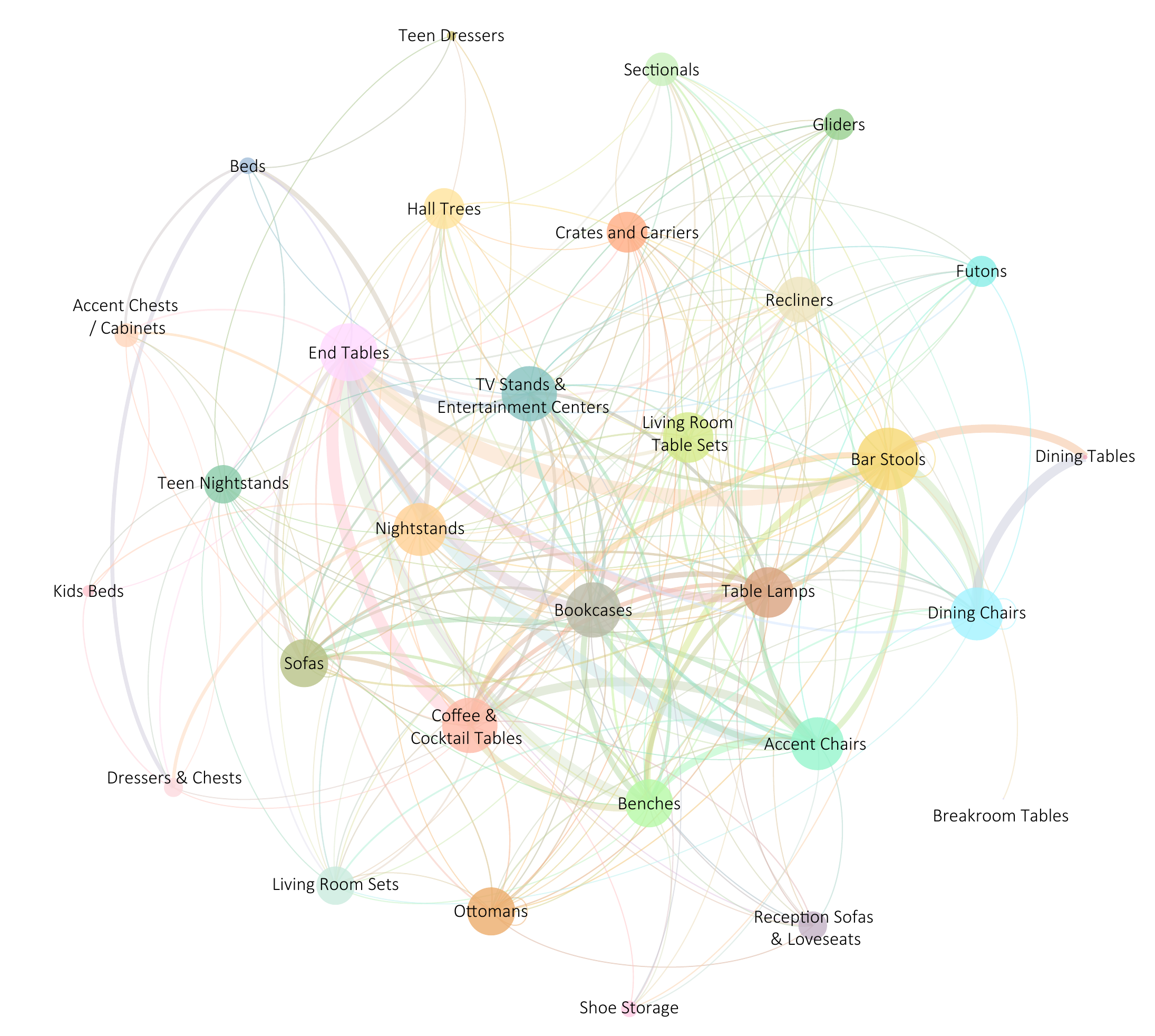}
\caption{Connections between the product groups.}
\label{fig:sku_classes_graph}    
\end{figure}

Figure~\ref{fig:sku_classes_graph} depicts a total of 30 nodes representing a group, where the size of each node relates to its degree. Edges drawn between product groups represent the cumulative weighting of the underlying product connections in the graph. The most common group is the End Table with $1,201$ products, followed by Bar Stools with $808$, and Accent Chairs with $784$. Crates and Carriers, Gliders, and Shoe Storage all having 10 nodes.

When viewing the consolidated network, trends among the product groups become apparent. First, we notice the product type itself heavily influences connections to other products as distinct groupings emerge, such as the strongly connected relationship between Bar Stools, Dining Tables, and Dining Chairs. As discussed previously, dataset noise may be a strong factor, since it is likely for similar products to be seen within product images, such as using a Dining Table as staging products for a Dining Chair image. Second, while the total number of End Tables exceeds all other product groups, the sum degree of the Bar Stools group is greater. This relationship may suggest a higher diversity in the styles and relationships a Bar Stool can be recommended for than End Tables. 

\begin{figure}[h]
    \begin{subfigure}[b]{0.24\textwidth}
        \includegraphics[width = \textwidth, page=9]{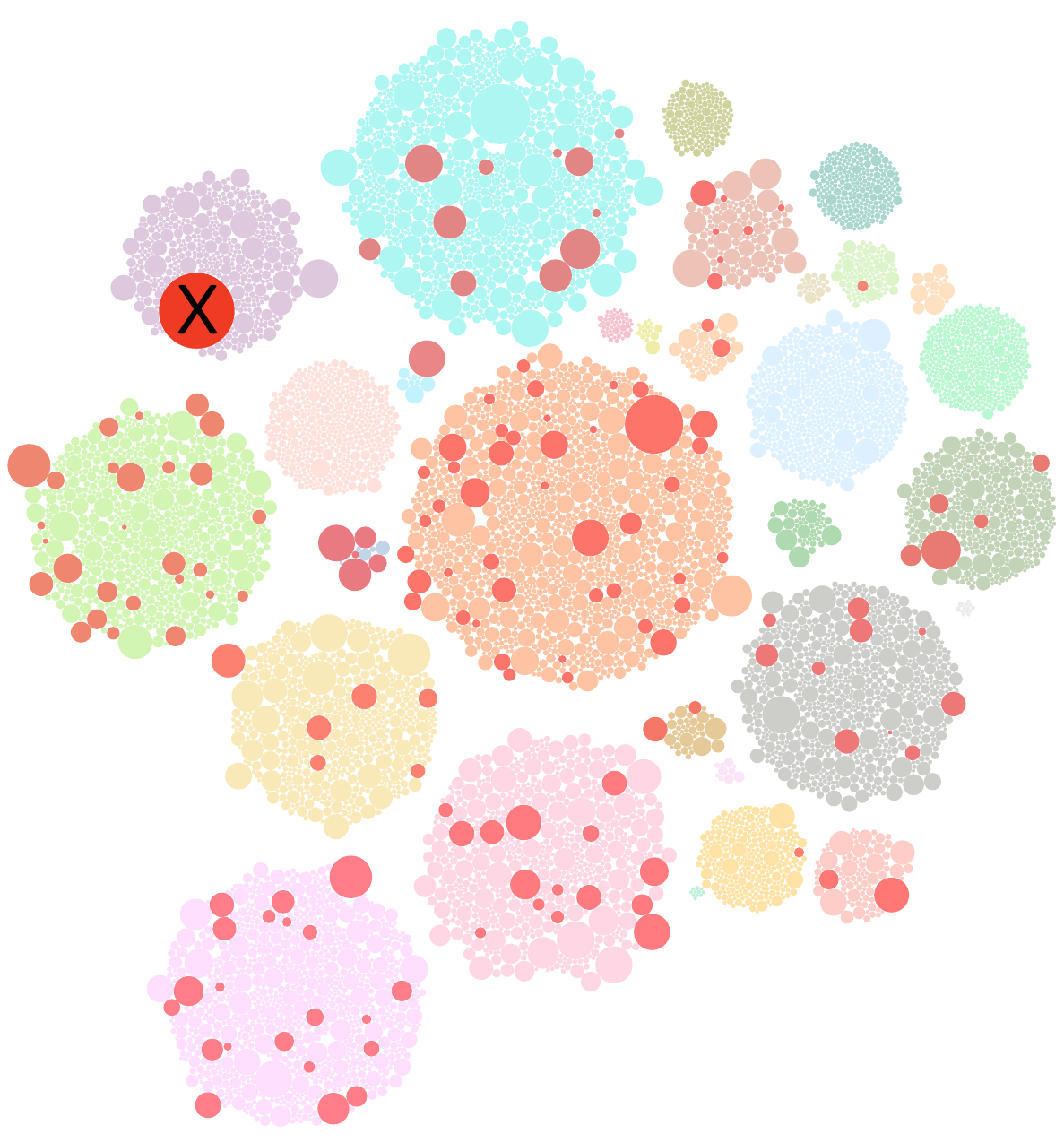}
        \caption{(One) TV Stands \& Entertainment Center}
        \label{subfig:tv_group_connections}  
    \end{subfigure}
    \begin{subfigure}[b]{0.24\textwidth}
        \includegraphics[width = \textwidth, page=11]{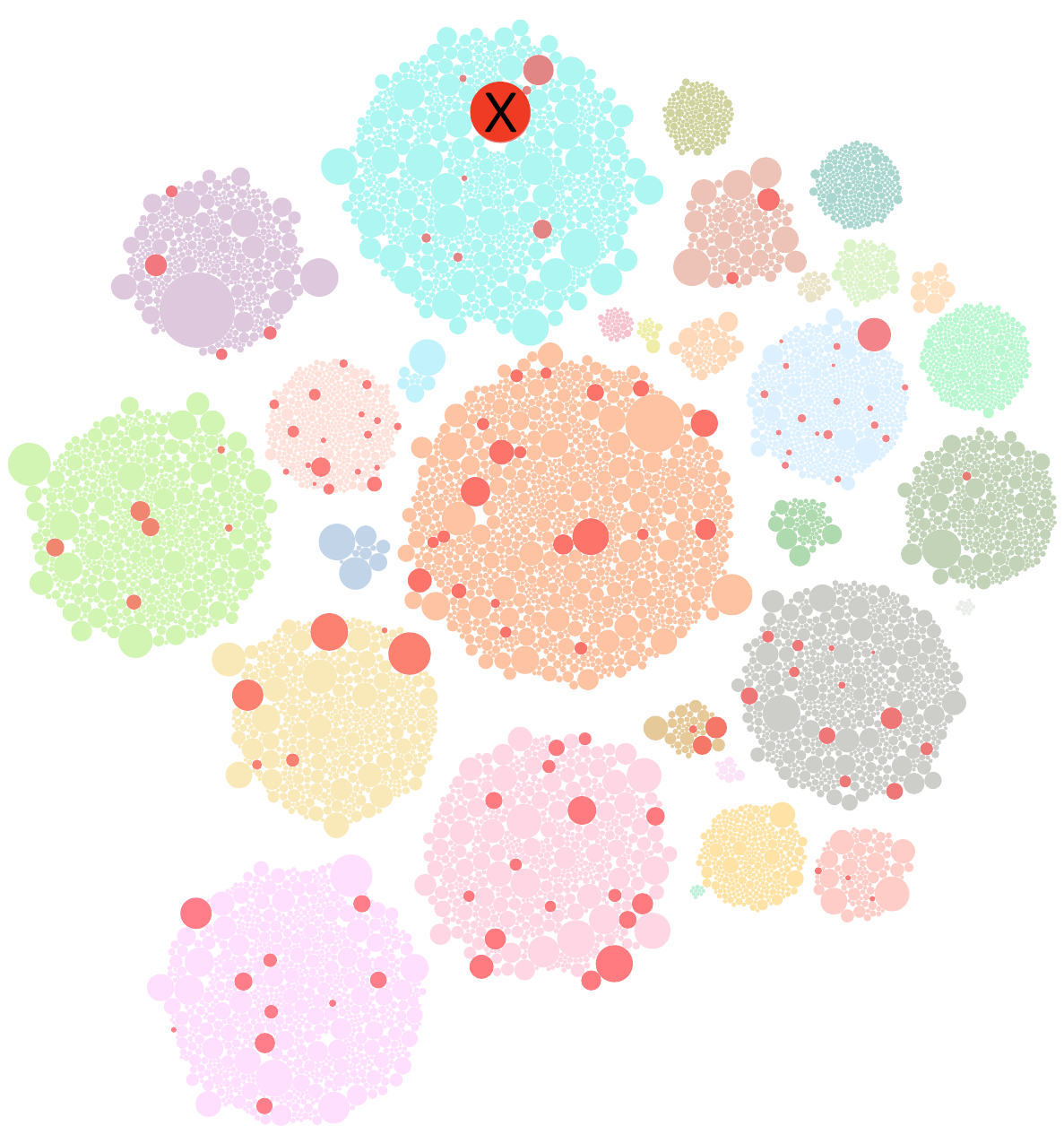}
        \caption{(One) Bar Stool}
        \label{subfig:bar_group_connections}  
    \end{subfigure}
     \begin{subfigure}[b]{0.24\textwidth}
        \includegraphics[width = \textwidth, page=12]{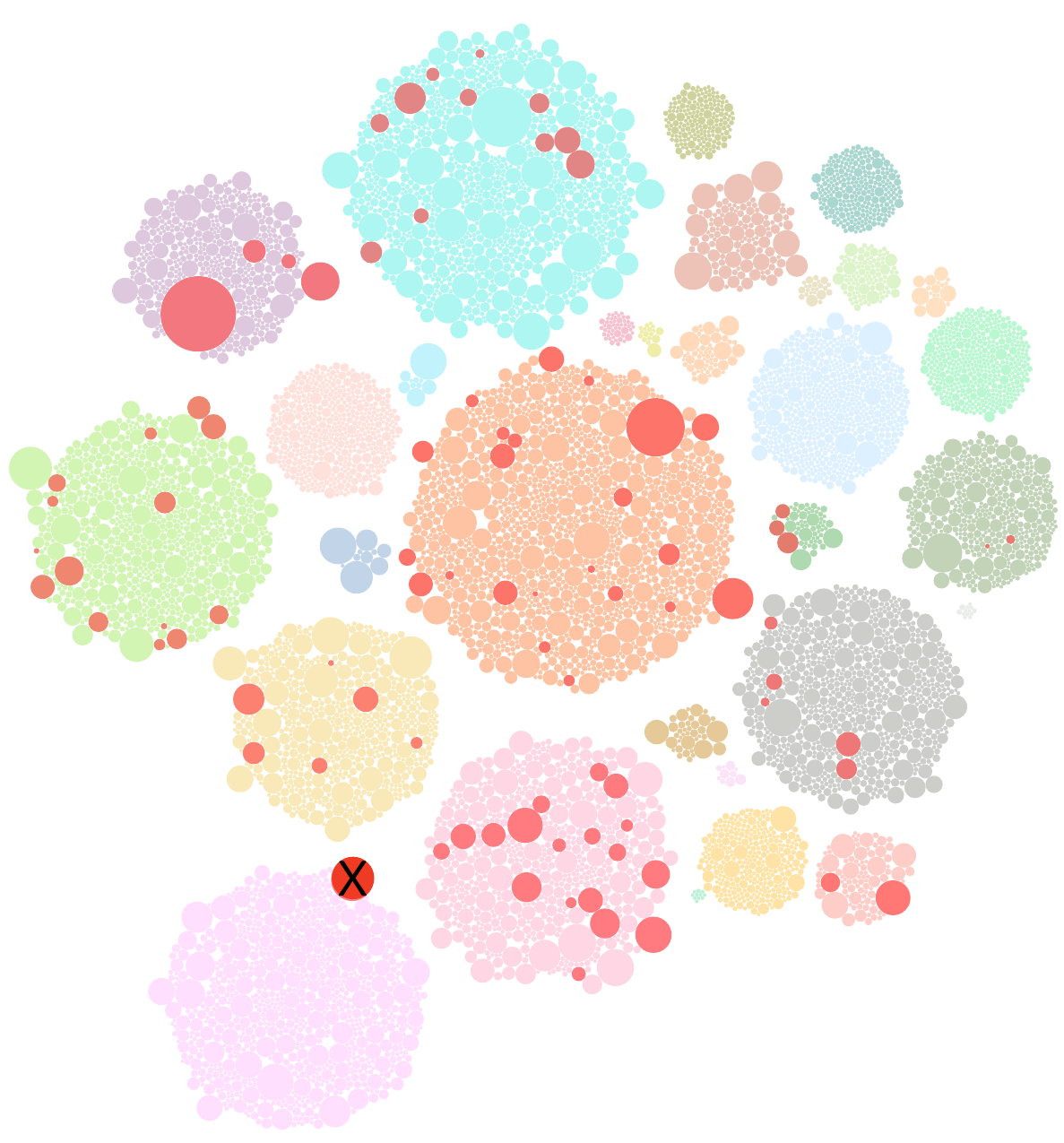}
        \caption{(One) Accent Chair}
        \label{subfig:accent_group_connections}  
    \end{subfigure}  
     \begin{subfigure}[b]{0.24\textwidth}
        \includegraphics[width = \textwidth, page=13]{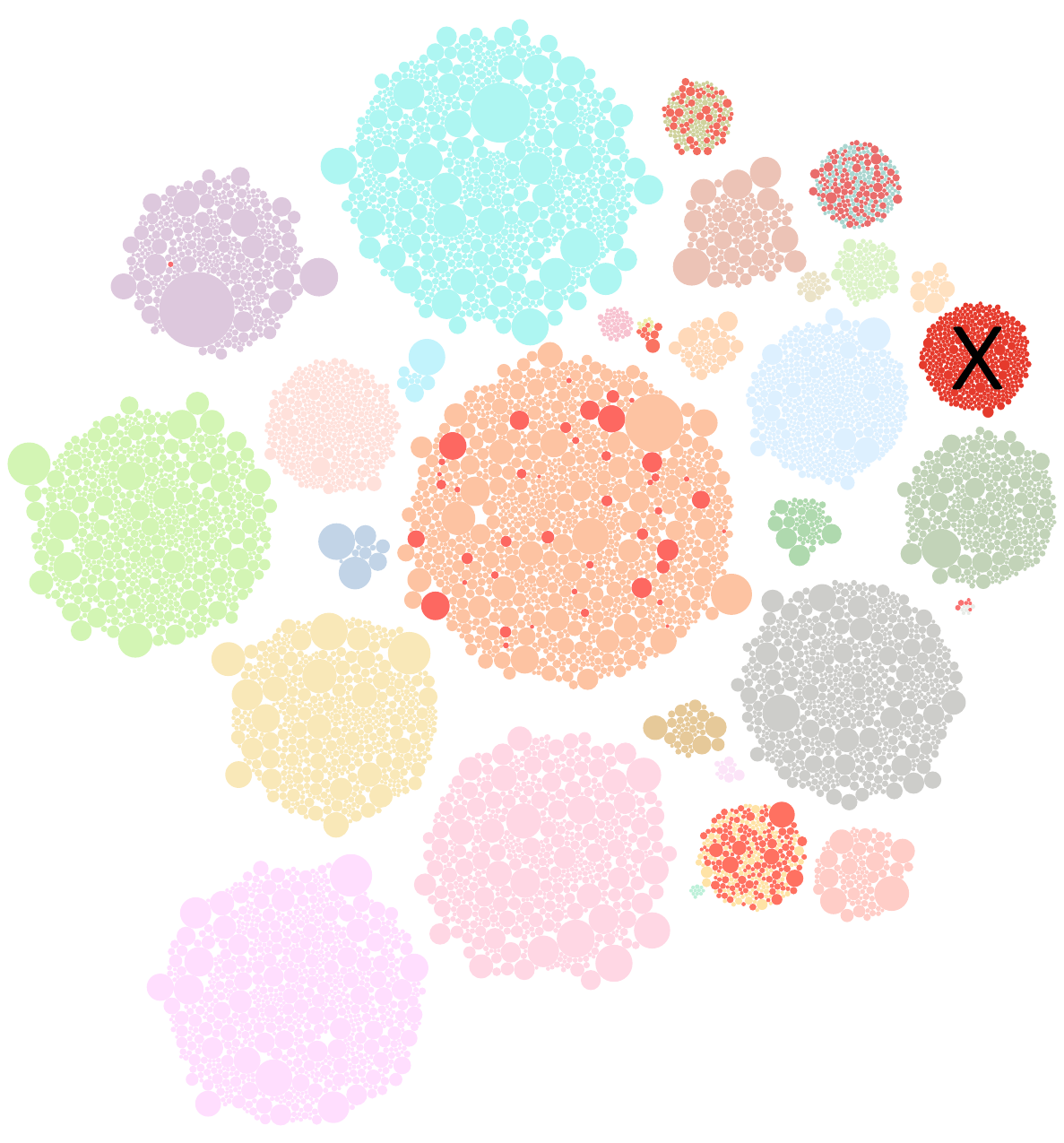}
        \caption{(All) Beds}
        \label{subfig:bed_group_connections}  
    \end{subfigure}   
\caption{Similar products of a selected product displayed in red. The groups are colored by the same scheme in Fig.~\ref{fig:sku_nodes_groups}. The most frequently recommended product is chosen as the target in the first three images (Fig.~\ref{subfig:tv_group_connections}-\ref{subfig:accent_group_connections}), while the last image (Fig.~\ref{subfig:bed_group_connections}) shows the recommendations from all of the products in the Bed group (the red cluster on right side of graph).}
\label{fig:product_group_connections}  
\end{figure}

To further understand group connectivity we explore relationships between the most connected products in Entertainment Centers, Bar Stools, and Accent Chairs (Fig.~\ref{fig:product_group_connections}). We find that Accent Chairs are commonly recommended for other categories (e.g., Entertainment Centers or Bar Stools), yet the most commonly recommended Accent Chair has no connection to other products within its own group (i.e., other Accent Chairs). This could be a result of weakly connected products within the same group being filtered out at the initial steps, or interestingly, a result of the style embeddings determining that the features of one Accent Chair do not strongly match those of other Accent Chairs. In this latter case, we interpret this relationship to mean that part of the \textit{style attributes} associated with an Accent Chair is the inherent Accent, in which combining this chair with others would be unlikely. Naturally the style itself can be similar between these chairs, yet as seen in the example data (Fig.~\ref{fig:product_examples}, images are a combination of the products within a scene or environment, possibly skewing the relationships to that of a more practical setting in which products exist rather than in isolation. In contrast to the wide-ranging connections between various product groups in Figures~\ref{subfig:tv_group_connections}-~\ref{subfig:accent_group_connections}, Figure~\ref{subfig:bed_group_connections} highlights (in red) the connections of all products in the Bed group (located on the right side). Such highlighting emphasizes product recommendations are limited to only a few groups: End Tables, Nightstands, Accent Chests / Cabinets, Dressers and Chests, and the smaller categories of Teen Dressers and Teen Nightstands. These results can be interpreted similarly to the Accent Chairs in which the training data lacked examples in which Beds appear in the same image as Bar Stools.

\begin{figure}[h]
\centering
    \begin{subfigure}[b]{0.49\textwidth}
        	\includegraphics[page=7, width=\textwidth]{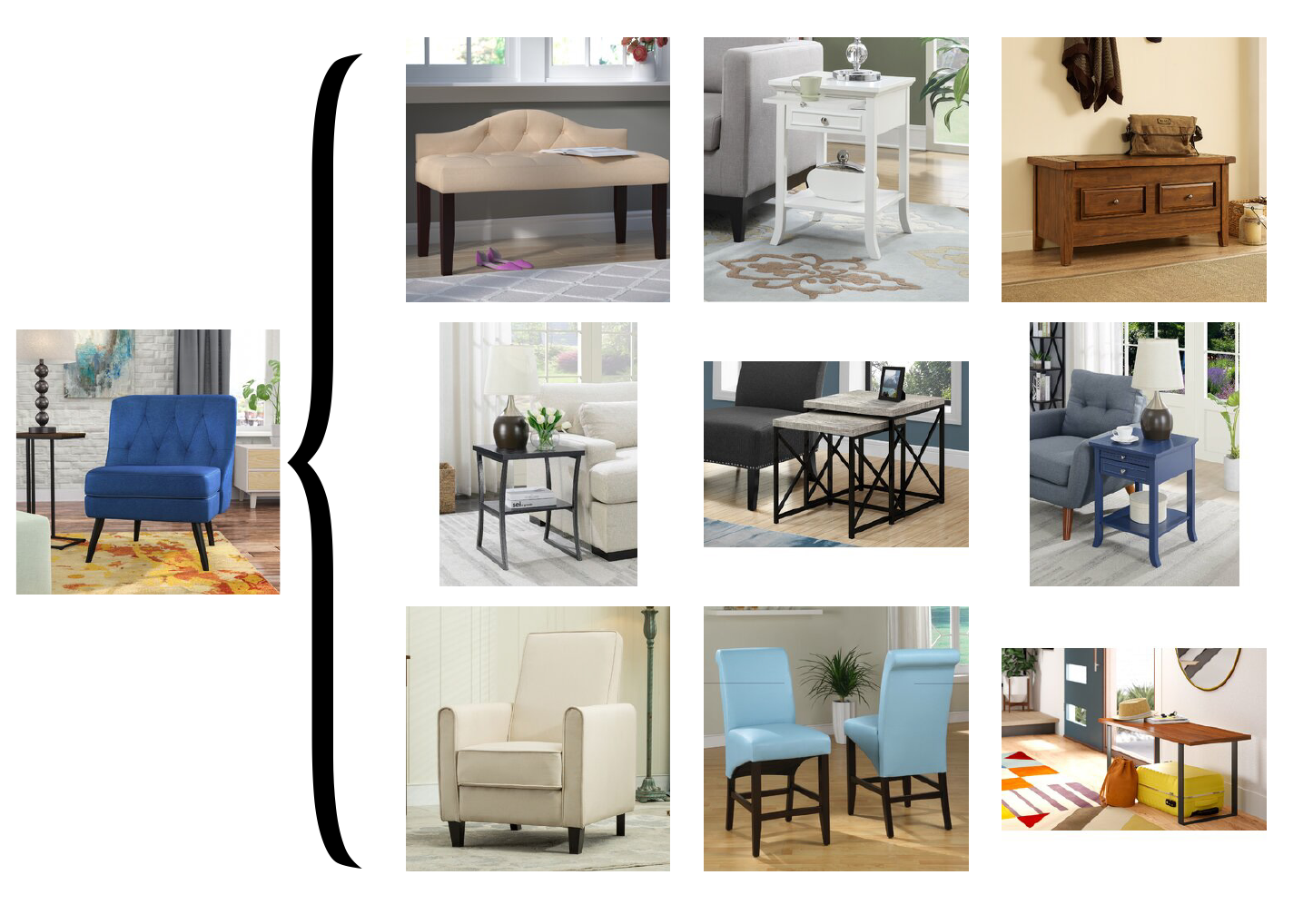}
        	\caption{Accent Chair}
        	\label{subfig:chair}
    \end{subfigure}
    \begin{subfigure}[b]{0.49\textwidth}
        	\includegraphics[page=8, width=\textwidth]{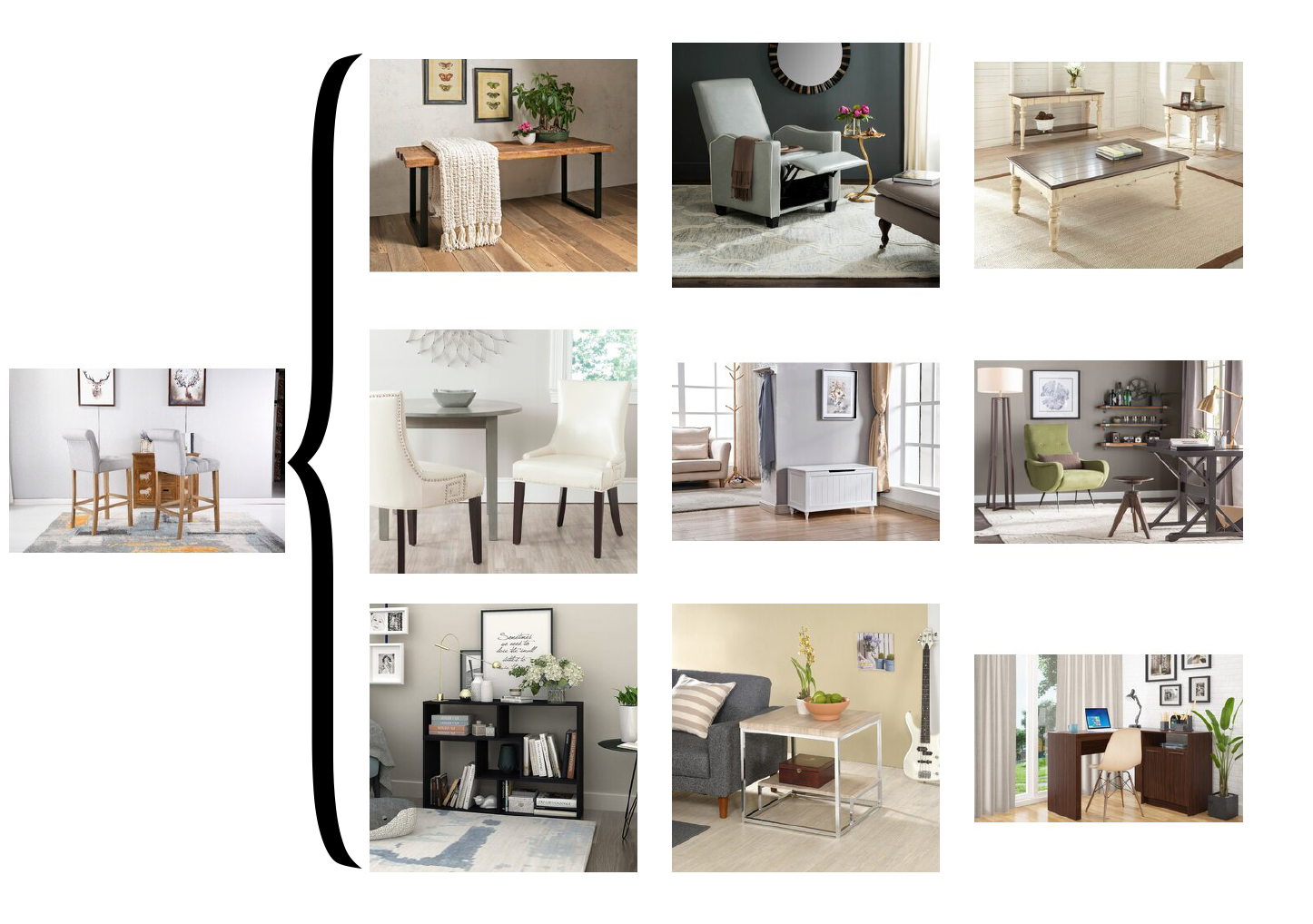}
        	\caption{Bar Stool}
        	\label{subfig:bar}
    \end{subfigure}
\caption{Nine most strongly connected recommendations for the most commonly recommended Accent Chair (Fig.~\ref{subfig:chair}) and Bar Stool(Fig.~\ref{subfig:bar}). }
\label{fig:top_rec_ex} 
\end{figure}

Keeping in mind an error rate of the machine learning algorithm on a more individual image scale, by exploring strong relationships to the most popular products in a group, we gain additional insights as to how a product may contain stylistic attributes determined by our neural network. One key point we found is that product groups found to be strongly connected to an input product generally do not seem to be dominated by color (Fig.~\ref{subfig:chair}). When viewing the strongly connected products in Figure~\ref{subfig:bar}, a theme of large, open, and box-like recommendations seem more common than in Figure~\ref{subfig:chair}. 

\section{Product Design Loop}
In the previous sections we describe how the neural network works and provide a discussion on some of the results. From this basis, we propose a method for integrating a product design workflow in a manner similar to that of the consumers interaction with the neural network on the e-commerce side. To support future work and continued use of our findings as algorithms in style analysis develop, the abstracted and generalized use case discussed in this section are independent of the speculative and data-exploratory methods we employed in the previous sections. 

\begin{figure}[h]
\sidecaption[t]
\includegraphics[width = .6\textwidth, page=10]{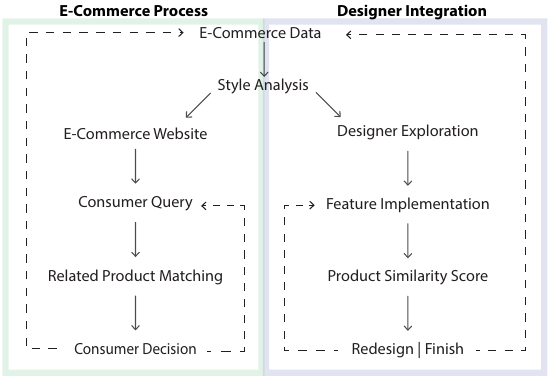}
\caption{(Left) E-Commerce process for showing related products to a consumer based on their query and shopping preferences. (Right) A mirrored example utilized by the designer to find product similarities of a proposed design.}
\label{fig:designer_in_loop}       
\end{figure}

Current E-Commerce systems use a variety of methods to propose alternative and additional products to a consumer; from reviews to collaborative filtering. In Figure~\ref{fig:designer_in_loop}, the left side illustrates this current practice with the right side demonstrating the \textit{Designer in-the-loop} process proposed here. In both cases, a list of possible products on the e-commerce site are run through an image-based neural network to create a graph of recommendations. On the consumer side, a query is given to the website on a particular product, which results in additional related product matches being displayed. The consumer can then choose from one of these options, after which such choice is updated at an e-commerce backend server (either as a purchase for reducing inventory, or stored for additional filtering methods). Alternatively, the consumer can perform a new query and repeat the process. For the designer, an initial exploration process begins with finding either gaps in the dataset where products are not being connected, or by finding strongly connected features between important/popular products. After implementing such desired features in a design, the rendered image (As the machine learning model is trained on both real and 3D rendered images),  can be fed to the algorithm which provides the designer with a product similarity score. From these connections, designer can either redesign and iterate, or finish the design, adding the finalized design to the e-commerce website listings. We propose that this process could enable automated brand cohesion checking as well.

\section{Discussion}

A main challenge and limitation in the final interpretation of our dataset relates to the uncertainty in the style-based analysis and image matching, as we rely on a subset of the output from the company-provided model. This has posed a few issues with extending our analysis and inferring additional aspects of product style. For one, there are many images that include the same or similar products. While we have a subset of the images labeled with SKUs within them, additional staging props or noise in this list can be meaningful in the recommendations. Similarly, while our data comes from one of the largest e-commerce companies, it is by no means a complete set of products and styles. Further implementations and additional product comparison data could provide insights to the addition of external styles.  For example, we are unsure of what would happen if a style not included in the training data is given to the model. Additionally, the subjective aspect of \textit{style} can be a hindrance to the machine learning model as culturally and over time the labels and attributes a human perceives as a certain style may change. However, with the increasing accuracy of machine learning models, or with more analysis options for a designer to regenerate style groupings to provide results they more easily understand, our proposed approach may provide a new process for integrating design and big-data analysis for e-commerce use cases.  

Wayfair, one of the largest e-commerce websites in the home and decor space, focuses on the huge selection, consistent experience, rich product information, frictionless tools, beautiful content, clear shipping and delivery, and great service to ensure that the experience of shopping for the home is easy and fun.  As a market-leading platform, beyond the customer’s needs, we focus on the supplier experience by providing the ability to tell a product’s story, get to market quickly, and support their business.  The capability to find a product design using qualities found in highly recommended products can fill a niche that may currently be underserved in the market and would enable the customers to quickly provide demand signals towards emerging market trends to suppliers.  Reducing the overall time in the feedback loop between suppliers and customers in the home and decor space will ensure suppliers have more complete information on market demand and will lead to a revolution in their planning and production process.

\begin{acknowledgement}
We thank Ilkay Yildiz and Hantian Liu for their preliminary work at Wayfair for the design and experimentation with the neural network style estimation model.
\end{acknowledgement}

\section*{Appendix}

\begin{figure}[h]
\sidecaption[t]
\includegraphics[width = .6\textwidth, page=4]{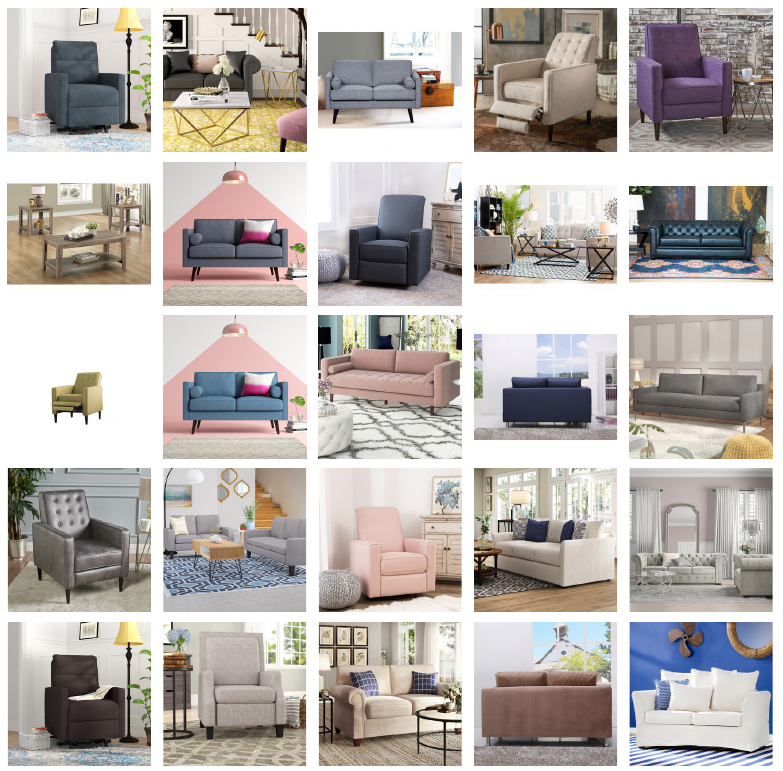}
\caption{Top-25 most frequently recommended images out of the 37,421 in our sample dataset, with the overall similarity threshold. 
Such image collection is independent of the products 
themselves, and exists across all of the original data, including low scoring similarities.}
\label{fig:most_freq_img_recs}     
\end{figure}

\addcontentsline{toc}{section}{Appendix}

\bibliographystyle{spmpsci}
\bibliography{culturalDNA_2020_arxiv}

\end{document}